\documentclass[wcp]{jmlr}

\usepackage{longtable}

\usepackage{booktabs}
\usepackage{natbib}

\usepackage{cite}
\usepackage{url}
\usepackage{amsmath,amssymb,amsfonts}
\usepackage{algorithmic}
\usepackage{graphicx}
\usepackage{textcomp}

\usepackage{courier}

\usepackage{caption}
\usepackage{subcaption}

\usepackage{array}
\newcolumntype{L}[1]{>{\raggedright\let\newline\\\arraybackslash\hspace{0pt}}m{#1}}
\newcolumntype{C}[1]{>{\centering\let\newline\\\arraybackslash\hspace{0pt}}m{#1}}
\newcolumntype{R}[1]{>{\raggedleft\let\newline\\\arraybackslash\hspace{0pt}}m{#1}}

\usepackage{tikz}
\usetikzlibrary{calc,fit,matrix,chains,positioning,decorations.pathreplacing,arrows}

\usepackage{xcolor}

\jmlrvolume{101}
\jmlryear{2019}
\jmlrworkshop{ACML 2019}

\title[Deep Learning with a Rethinking Structure for Multi-label Classification]{Deep Learning with a Rethinking Structure \\ for Multi-label Classification}

  \author{\Name{Yao-Yuan Yang} \Email{b01902066@ntu.edu.tw}\\
   \Name{Yi-An Lin} \Email{andylin514@gmail.com}\\
   \Name{Hong-Min Chu} \Email{r04922031@csie.ntu.edu.tw}\\
   \Name{Hsuan-Tien Lin} \Email{htlin@csie.ntu.edu.tw}\\
   \addr Department of Computer Science and Information Engineering, National Taiwan University
 }

\begin{document}

\maketitle

\begin{abstract}
Multi-label classification (MLC) is an important class of machine learning problems that come with a wide spectrum of applications, each demanding a possibly different evaluation criterion. When solving the MLC problems, we generally expect the learning algorithm to take the hidden correlation of the labels into account to improve the prediction performance. Extracting the hidden correlation is generally a challenging task. In this work, we propose a novel deep learning framework to better extract the hidden correlation with the help of the memory structure within recurrent neural networks. The memory stores the temporary guesses on the labels and effectively allows the framework to rethink about the goodness and correlation of the guesses before making the final prediction. Furthermore, the rethinking process makes it easy to adapt to different evaluation criteria to match real-world application needs. In particular, the framework can be trained in an end-to-end style with respect to any given MLC evaluation criteria. The end-to-end design can be seamlessly combined with other deep learning techniques to conquer challenging MLC problems like image tagging. Experimental results across many real-world data sets justify that the rethinking framework indeed improves MLC performance across different evaluation criteria and leads to superior performance over state-of-the-art MLC algorithms.
\end{abstract}

\section{Introduction}

Human beings master our skills for a given problem by working on and thinking
through the same problem over and over again.
When a difficult problem is given to us, multiple attempts would have
gone through our mind to simulate different possibilities.
During this period, our understanding to the problem gets deeper, which in term 
allows us to propose a better solution in the end.
The deeper understanding comes from a piece of consolidated knowledge within our memory,
which records how we build up the problem context with processing and
predicting during the ``rethinking'' attempts.
The human-rethinking model above inspires us to design a novel deep learning
model for machine-rethinking, which is equipped with a memory structure
to better solve the multi-label classification (MLC) problem.

The MLC problem aims to attach multiple relevant labels to an input instance
simultaneously, and matches various application scenarios, such as tagging
songs with a subset of emotions \citep{DBLP:conf/ismir/TrohidisTKV08} or
labeling images with objects \citep{wang2016cnn}.
Those MLC applications typically come with an important property called label correlation~\citep{cheng2010bayes,huang2012multi}.
For instance, when tagging songs with emotions, ``angry'' is negatively correlated with ``happy''; when labeling images, the existence of a desktop computer probably indicates the co-existence of a keyboard and a mouse.
Many existing MLC works implicitly or explicitly take label
correlation into account to better solve MLC problems~\citep{cheng2010bayes}. 

Label correlation is also known to be important for human when solving MLC problems \citep{bar2004visual}.
For instance, when solving an image labeling task upon entering a new room,
we might notice some more obvious objects like sofa, dining
table and wooden floor at the first glance.
Such a combination of objects hints us of a living room, which helps us better
recognize the ``geese'' on the sofa to be stuffed animals instead of real ones.
The recognition route from the sofa to the living room to stuffed animals require rethinking
about the correlation of the predictions step by step.
Our proposed machine-rethinking model mimics this human-rethinking process to digest label correlation and solve MLC problems more accurately.

Next, we introduce some representative MLC algorithms
before connecting them to our proposed machine-rethinking model.
Binary relevance (BR)~\citep{tsoumakas2009mining} is
a baseline MLC algorithm that does not consider label correlation.
For each label, BR learns a binary classifier to predict the label's relevance independently.
Classifier chain (CC)~\citep{read2009classifier} extends
BR by taking some label correlation into account.
CC links the binary classifiers as a chain and feeds the predictions of the earlier classifiers as
features to the latter classifiers. The 
latter classifiers can thus utilize (the correlation to) the earlier predictions to form better predictions.

The design of CC can be viewed as a memory mechanism that
stores the label predictions of the earlier classifiers.
Convolutional neural network recurrent neural network~(CNN-RNN)~\citep{wang2016cnn}
and order-free RNN with Visual Attention~(Att-RNN)~\citep{DBLP:journals/corr/ChenCYW17}
algorithms extend CC by replacing the mechanism with a more
sophisticated memory-based model---recurrent neural network (RNN).
By adopting different variations of RNN~\citep{hochreiter1997long,cho2014learning}, the memory can store more sophisticated
concepts beyond earlier predictions.
In addition, adopting RNN allows the algorithms to solve tasks like image labeling more effectively via end-to-end training with
other deep learning architectures (e.g., convolutional neural network in CNN-RNN).



The CC-family algorithms above for utilizing label correlation are
reported to achieve better performance than
BR~\citep{read2009classifier,wang2016cnn}. Nevertheless,
given that the predictions happen sequentially within a chain, those algorithms
generally suffer from the issue of label ordering. 
In particular, classifiers in different positions of the chain receive different levels of information.
The last classifier predicts with all information from other
classifiers while the first classifier label predicts with no other information.
Att-RNN addresses this issue with beam search to approximate
the optimal ordering of the labels, and dynamic programming based classifier chain (CC-DP)~\citep{liu2015optimality}
searches for the optimal ordering with dynamic programming.
Both Att-RNN and CC-DP can be time-consuming when searching for the optimal ordering,
and even after identifying a good ordering, the label correlation information
is still not shared equally during the prediction process.


Our proposed deep learning model, called RethinkNet, tackles the label ordering issue by viewing CC differently. By considering CC-family algorithms as a rethinking model based on the partial predictions from earlier classifiers, we propose to \textit{fully} memorize the temporary predictions from \textit{all} classifiers during the rethinking process. That is, instead
of forming a chain of binary classifiers, we form a chain of \textit{multi-label} classifiers as a sequence of rethinking.
RethinkNet learns to form preliminary guesses in earlier classifiers of the chain, store those guesses in the memory and then correct those guesses in latter classifiers with label correlation.
Similar to CNN-RNN and Att-RNN, RethinkNet adopts RNN for making memory-based sequential prediction. We design a global memory for RethinkNet to
store the information about label correlation, and the global memory
allows all classifiers to share the same information without suffering
from the label ordering issue.

Another advantage of RethinkNet is to tackle an important real-world need of
Cost-Sensitive Multi-Label Classification (CSMLC) \citep{CL2014b}.
In particular, different MLC applications often
require different evaluation criteria. To be widely useful for a broad spectrum of applications, it is thus important to design
CSMLC algorithms, which takes the criteria (cost) into account during learning and can thus adapt to different costs easily.
State-of-the-art CSMLC algorithms include condensed filter tree (CFT)
\citep{CL2014b} and probabilistic classifier chain (PCC)
\citep{cheng2010bayes}.
PCC extends CC to CSMLC by making Bayes optimal predictions according
to the criterion.
CFT is also extended from CC, but achieves cost-sensitivity by converting the criterion to importance weights when training each binary classifier within CC.
The conversion step in CFT generally requires knowing the predictions of all classifiers, which has readily been stored within the memory or RethinkNet. Thus,
RethinkNet can be easily combined with the importance-weighting idea within CFT to achieve cost-sensitivity.

Extensive experiments
across real-world data sets validate that RethinkNet indeed improves MLC performance across different evaluation criteria and is superior to state-of-the-art MLC and CSMLC algorithms. Furthermore, for image labeling, experimental results demonstrate that RethinkNet outperforms both CNN-RNN and Att-RNN. The results
justify the usefulness of RethinkNet.

The paper is organized as follows. Section~\ref{problem-definition} sets up
the problem and introduces concurrent RNN models.
Section~\ref{proposed-model} illustrates the proposed RethinkNet framework.
Section~\ref{experiments} contains extensive experimental results to
demonstrate the benefits of RethinkNet. Finally, Section~\ref{conclusion}
concludes our findings.

\section{Preliminary}\label{problem-definition}

In the multi-label classification (MLC) problem, the goal is to attach multiple
labels to a feature vector $\mathbf{x} \in \mathcal{X} \subseteq \mathbb{R}^d$.
Let there be a total of $K$ labels.
The labels are represented by a label vector $\mathbf{y} \in \mathcal{Y}
\subseteq \{0, 1\}^K$, where the $k$-th bit $\mathbf{y}[k] = 1$ if and only
if the $k$-th label is relevant to $\mathbf{x}$.
We call $\mathcal{X}$ and $\mathcal{Y}$ the feature space and the label space, respectively.

During training, a MLC algorithm takes the training data set
$\mathcal{D} = \{(\mathbf{x}_{n}, \mathbf{y}_{n})\}_{n=1}^N$ that
contains $N$ examples with $\mathbf{x}_{n} \in \mathcal{X}$ and $\mathbf{y}_{n} \in \mathcal{Y}$
to learn a classifier $f\colon \mathcal{X}~\to~\mathcal{Y}$. The classifier
maps a feature vector $\mathbf{x} \in \mathcal{X}$ to its predicted label vector in $\mathcal{Y}$.
For testing, $\bar{N}$ test examples $\{(\mathbf{\bar{x}}_n, \mathbf{\bar{y}}_n)\}_{n=1}^{\bar{N}}$ are drawn from
the same distribution that generated the training data set $\mathcal{D}$.
The goal of an MLC algorithm is to learn a classifier $f$ such that
the predicted vectors
$\{\mathbf{\hat{y}}_{n}\}_{n=1}^{\bar{N}} = \{f(\mathbf{\bar{x}}_{n})\}_{n=1}^{\bar{N}}$
such that $\{\mathbf{\hat{y}}_{n}\}_{n=1}^{\bar{N}}$ are close to the ground truth vectors $\{\mathbf{\bar{y}}_{n}\}_{n=1}^{\bar{N}}$.

The closeness between label vectors is measured by
the evaluation criteria.
Different applications require possibly different evaluation criteria, which calls
for a more general setup called cost-sensitive multi-label classification
(CSMLC).
In this work, we follow the setting from previous works \citep{CL2014b,cheng2010bayes} and
consider a specific family of evaluation criteria.
This family of criteria measures the closeness between a single predicted vector
$\mathbf{\hat{y}}$ and a single ground truth vector $\mathbf{y}$.
To be more specific, these criteria can be written as a cost function
$C\colon \mathcal{Y} \times \mathcal{Y} \to \mathbb{R}$, where $C(\mathbf{y},
\mathbf{\hat{y}})$ represents the cost (difference) of predicting~$\mathbf{y}$
as $\mathbf{\hat{y}}$.
This way, classifier $f$ can be evaluated by the average cost
$\frac{1}{\bar{N}} \sum^{\bar{N}}_{n=1} C(\mathbf{\bar{y}}_{n}, \mathbf{\hat{y}}_{n})$ on the test examples.

In CSMLC setting, we assume the criterion for evaluation to be known before
training.
That is, CSMLC algorithms can learn $f$ with both the training data
set~$\mathcal{D}$ and the cost function~$C$, and should be able to adapt to
different $C$ easily.
By using this additional cost information, 
CSMLC algorithms aim at minimizing the expected cost on the test data set
$\mathbb{E}_{(\mathbf{x}, \mathbf{y})\sim\mathcal{D}}[C(\mathbf{y}, f(\mathbf{x}))]$.
Common cost functions are listed as follows.
Note that some common `cost' functions use higher output to represent
a better prediction---we call those \textit{score} functions to differentiate
them from usual cost (\textit{loss}) functions that use lower output to represent a better prediction.

\begin{itemize}
  \setlength\itemsep{.2em}
  \item Hamming loss:
$C_H(\mathbf{y}, \mathbf{\hat{y}})= \frac{1}{K} \sum_{k=1}^K
[\![\mathbf{y}[k] \neq \mathbf{\hat{y}}[k]]\!]$
where $[\![ \cdot ]\!]$ is the indicator function.

  \item F1 score:
$C_F(\mathbf{y}, \mathbf{\hat{y}}) = \frac{2\| \mathbf{y} \cap
\mathbf{\hat{y}}\|_1}{\|\mathbf{y}\|_1 + \|\mathbf{\hat{y}}\|_1}$,
where $\| \mathbf{y} \|_1$ actually equals the number of $1$'ss in label vector $\mathbf{y}$.

  \item Accuracy score:
$C_A(\mathbf{y}, \mathbf{\hat{y}}) = \frac{\|\mathbf{y} \cap \mathbf{\hat{y}}\|_1}{\|\mathbf{y} \cup \mathbf{\hat{y}}\|_1}$

  \item Rank loss:
$C_R(\mathbf{y}, \mathbf{\hat{y}}) = \sum_{\mathbf{y}[i]>\mathbf{y}[j]}\left([\![\mathbf{\hat{y}}[i]<\mathbf{\hat{y}}[j]]\!] +
\frac{1}{2}[\![\mathbf{\hat{y}}[i]=\mathbf{\hat{y}}[j]]\!]\right)$

\end{itemize}

\subsection{Related Work}

There are many different families of MLC algorithms.
In this work, we focuses on the chain-based algorithms, which make
predictions label by label sequentially and each prediction take previously-predicted labels as inputs.

Classifier chain~(CC)~\citep{read2009classifier} is the most classic
algorithm in the chain-based family.
CC learns a sub-classifier per label. In particular,
CC predicts the label one by one, the prediction of previous labels are fed
to the next sub-classifier to predict the next label.
CC allows sub-classifiers to utilize label correlation by building
correlations between sub-classifiers.

However, deciding the label ordering for CC can be difficult and 
the label ordering is crucial to the performance
\citep{read2009classifier,read2014efficient,goncalves2013genetic,liu2015optimality}.
The sub-classifier in the later part of the chain can receive more
information from other sub-classifiers while others receive less.
Algorithms have been developed to solve the label ordering problem by using
different ways to search for a better ordering of the labels.
These algorithms include one that uses monte carlo methods
\citep{read2014efficient} and genetic algorithm \citep{goncalves2013genetic}.
A recent work called
dynamic programming based classifier chain (CC-DP)~\citep{liu2015optimality} is
proposed that by using dynamic programming.
However, the time complexity for CC-DP is still as large as $O(K^3Nd)$ and
the derivation is limited using support vector machine~(SVM) as
sub-classifier.

For the CSMLC setup, there are two algorithms developed based on CC, the
probabilistic classifier chain (PCC)~\citep{cheng2010bayes}
and condense filter tree~(CFT)~\citep{CL2014b}.
PCC learns a CC classifier during training.
During testing, PCC make a bayes optimal decision with respect to the given
cost function.
This step of making inference can be time consuming, thus efficient inference
rule for each cost function are needed to be derived individually.
Efficient inference rule for F1 score and Rank loss are derived in
\citep{dembczynski2012consistent,dembczynski2011exact}.
Albeit, there is no known inference rule for Accuracy score.
For CFT, it transforms the cost information into instance weight.
Through a multi-step training process, CFT gradually fine-tune the weight assigned
and itself cost-sensitive during training.
CFT does not require the inference rule to be derived for each cost function,
but the multi-step training process is still time consuming.
Also CFT is not able to be combined with deep learning architectures for
image or sound data sets.

CC can be interpreted as a deep learning architecture \citep{read2014deep}.
Each layer of the deep learning architecture predicts a label and passes the prediction
to the next layer.
By turning the idea of building correlations between sub-classifiers into
maintaining a memory between sub-classifiers,
the convolutional neural network recurrent neural network~(CNN-RNN)~\citep{wang2016cnn}
algorithm further adapt recurrent neural network~(RNN) to generalize the deep
learning architecture for CC.
CNN-RNN treats each prediction of the label as a time step in the RNN.
CNN-RNN also demonstrated that with this architecture, they are able to
incorporate with convolutional neural networks~(CNN) and produces
experimental results that outperforms traditional MLC algorithms.
The order-free RNN with Visual Attention~(Att-RNN)~\citep{DBLP:journals/corr/ChenCYW17}
algorithms is an improvement over CNN-RNN.
Att-RNN incorporated the attention model \citep{xu2015show} to enhance their
performance and interoperability.
Also, Att-RNN uses the beam search method to search for a better ordering
of the label.
But both Att-RNN and CNN-RNN are not cost-sensitive, thus they are unable to
utilize the cost information.

To summarize, there are three key aspects in developing a MLC algorithm.
Whether the algorithm is able to effectively utilize the label correlation information,
whether the algorithm is able to consider the cost information and
whether the algorithm is extendable to deep learning structures for modern application.
In terms of utilizing label correlation information, current chain based MLC
algorithms has to solve the label ordering due to the sequential nature of
the chain.
The first label and the last label in the chain are destined to receive
different amount of information from other labels.
Chain based algorithms are generally made extendable to other deep learning
architectures by adapting RNN.
However, there is currently no MLC algorithm in the chain based family that are
designed both considering cost information as well as being extendable with
other deep learning architectures.
In the next section, we will introduce the RNN to understand how it is
designed.

\subsection{Recurrent Neural Network (RNN)}

Recurrent Neural Network~(RNN) is a class of neural network model that are
designed to solve sequence prediction problem.
In sequence prediction problem, let there be $B$ iterations.
There is an output for each iteration.
RNN uses memory to pass information from one iteration to the next iteration.
RNN learns two transformations and all iterations shares these two transformations.
The feature transformation $\mathbf{U}(\cdot)$ takes in the feature vector
and transforms it to an output space.
The memory transformation $\mathbf{W}(\cdot)$ takes in the output from the
previous iteration and transform it to the same output space as the output of $\mathbf{U}$.
For $1 \le i \le B$, we use $\mathbf{x}^{(i)}$ to represent the feature
vector for the $i$-th iteration, and use $\mathbf{o}^{(i)}$ to represent its
output vector.
Formally, for $2 \le i \le B$, and let $\sigma(\cdot)$ be the activation
unction, the RNN model can be written as
$\mathbf{o}^{(1)} = \sigma(U(\mathbf{x}^{(1)})), \quad
\mathbf{o}^{(i)} = \sigma(U(\mathbf{x}^{(i)}) + W(\mathbf{o}^{(i-1)}))$.

The basic variation of RNN is called simple RNN
(SRN)~\citep{elman1990finding,jordan1997serial}.
SRN assumes $\mathbf{W}$ and $\mathbf{U}$ to be linear transformation.
SRN is able to link information between iterations, but it can be hard to train
due to the decay of gradient~\citep{hochreiter2001gradient}.
Several variations of RNN had been proposed to solve this problem.
Long short term memory (LSTM)~\citep{hochreiter1997long} and 
gated recurrent unit (GRU)~\citep{cho2014learning} solve this problem
by redesigning the neural network architecture.
Iterative RNN (IRNN)~\citep{le2015simple} proposed that the problem can be
solved by different initialization of the memory matrix in SRN.

In sum, RNN provides a foundation for sharing information
between a sequence of label predictions using memory.
In the next section, we will demonstrate how we utilize RNN to develope a
novel chain based MLC algorithms that addresses the three key aspects for MLC
algorithm mentioned in the previous subsection.

\section{Proposed Model}\label{proposed-model}

The idea of improving prediction result by iteratively polishing the
prediction is the ``rethinking'' process.
This process can be taken as a sequence prediction problem and RethinkNet
adopts RNN to model this process.

\begin{figure}[ht!]
\centering
\scalebox{0.8}{
\begin{tikzpicture}[shorten >=1pt,->,draw=black!50, node distance=\layersep]
    \tikzstyle{every pin edge}=[<-,shorten <=1pt]
    \tikzstyle{neuron}=[circle,fill=black!25,minimum size=30pt,inner sep=0pt]
    \tikzstyle{input neuron}=[neuron, fill=green!50];
    \tikzstyle{output neuron}=[neuron, fill=red!50];
    \tikzstyle{hidden neuron}=[neuron, fill=blue!50];
    \tikzstyle{dense}=[rectangle,fill=black!25,minimum width=25pt,minimum height=5cm,inner sep=0pt];
    \tikzstyle{annot}=[text width=4em, text centered]

    \node[] (x) at (-2cm, 3cm) {$\mathbf{x}$};

    \node[neuron] (r1) at (0, 5cm) {$t=1$};
    \node[neuron] (r2) at (0, 3cm) {$t=2$};
    \node[neuron] (r3) at (0, 1cm) {$t=3$};

    \node[dense] (emb) at (2.3cm, 3cm) {};

    \draw[->] (x) -- node[above] {} (r1);
    \draw[->] (x) -- node[above] {} (r2);
    \draw[->] (x) -- node[above] {} (r3);

    \draw[->] (r1) -- node[above] {$\mathbf{\hat{o}}^{(1)}$} (2.3cm-12.5pt, 5cm);
    \draw[->] (r2) -- node[above] {$\mathbf{\hat{o}}^{(2)}$} (2.3cm-12.5pt, 3cm);
    \draw[->] (r3) -- node[above] {$\mathbf{\hat{o}}^{(3)}$} (2.3cm-12.5pt, 1cm);

    \draw[->] (2.3cm+12.5pt, 5cm) -- node[above] {$\mathbf{\hat{y}}^{(1)}$} +(1.5cm,0);
    \draw[->] (2.3cm+12.5pt, 3cm) -- node[above] {$\mathbf{\hat{y}}^{(2)}$} +(1.5cm,0);
    \draw[->] (2.3cm+12.5pt, 1cm) -- node[above] {$\mathbf{\hat{y}}^{(3)}$} +(1.5cm,0);

    \draw[->] (r1) -- node[] {} (r2);
    \draw[->] (r2) -- node[out=30] {} (r3);

    \node[annot] (hf) at (-2.2cm, 1pt) {Feature vector};
    \node[annot] (hr) {RNN layer};
    \node[annot] (hd) at (2.3cm, 0) {Dense layer}  ;
\end{tikzpicture}
}
\caption{The architecture of the proposed RethinkNet model.} \label{fig:rethinknet}
\end{figure}
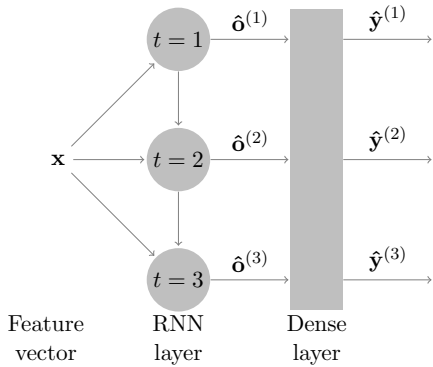

Figure \ref{fig:rethinknet} illustrates how RethinkNet is designed.
RethinkNet is composed of an RNN layer and a dense (fully connected) layer.
The dense layer learns a label embedding to transform the output of RNN
layer to label vector.
The RNN layer is used to model the ``rethinking'' process.
The RNN layer goes through a total of $B$ iterations.
All iterations in RNN share the same feature vector since they are solving
the same MLC problem.
The output of the RNN layer at $t$-th iteration is $\mathbf{\hat{o}}^{(t)}$, which
represents the embedding of the label vector $\mathbf{\hat{y}}^{(t)}$.
Each $\mathbf{\hat{o}}^{(t)}$ is passed down to $(t+1)$-th iteration in the
RNN layer.

Each iteration in RNN layer represents to a rethink iteration.
In the first iteration, RethinkNet makes a prediction base on the feature
vector alone, which targets at labels that are easier to identify.
The first prediction $\mathbf{\hat{y}}^{(1)}$ is similar to BR, which
predicts each label independently without the information of other labels.
From the second iteration, RethinkNet begins to use the result from the
previous iteration to make better predictions $\mathbf{\hat{y}}^{(2)} \ldots
\mathbf{\hat{y}}^{(B)}$.
$\mathbf{\hat{y}}^{(B)}$ is taken as the final prediction $\mathbf{\hat{y}}$.
As RethinkNet polishes the prediction, difficult labels would eventually be
labeled more accurately.

\subsection{Modeling Label Correlation}
RethinkNet models label correlation in the memory of the RNN layer.
To simplify the illustration, we assume that the activation function $\sigma(\cdot)$
is sigmoid function and the dense layer is an identity transformation.
SRN is used in the RNN layer because other forms of RNN share similar property since they are originated from SRN.
In SRN, the memory and feature transformations are represented as matrices
$\mathbf{W} \in R^{K \times K}$ and $\mathbf{U} \in R^{K \times d}$ respectively.
The RNN layer output $\mathbf{\hat{o}}^{(t)}$ will be a label vector with length $K$.

Under the setting, the predicted label vector is
$\mathbf{\hat{y}}^{(t)} = \mathbf{\hat{o}}^{(t)} = \sigma(\mathbf{U} \mathbf{x} + \mathbf{W} \mathbf{\hat{o}}^{(t-1)})$.
This equation can be separated into two parts, the feature term
$\mathbf{U} \mathbf{x}$, which makes the prediction like BR, and the memory
term $\mathbf{W} \mathbf{\hat{o}}^{(t-1)}$, which transforms the
previous prediction to the current label vector space.
This memory transformation serves as the model for label correlation.
$\mathbf{W}[i,j]$ represents $i$-th row and $j$-th column of $\mathbf{W}$ and
it represents the correlation between $i$-th and $j$-th label.
The prediction of $j$-th label is the combination of 
$(\mathbf{U} \mathbf{x})[j]$ and
$\mathbf{\hat{o}}^{(t)}[j] = \sum_{i=1}^{K} \mathbf{\hat{o}}^{(t-1)}[i] * \mathbf{W}[i,j]$.
If we predict $\mathbf{\hat{o}}^{(t-1)}[i]$ as relevant at $(t-1)$-th iteration and $\mathbf{W}[i,j]$ is high, it indicates that the $j$-th label is more
likely to be relevant.
If $\mathbf{W}[i,j]$ is negative, this indicates that the $i$-th label
and $j$-th label may be negatively correlated.

\begin{figure}[ht!]
  \centering
  \subfigure[memory transform]{\includegraphics[width=.3\textwidth]{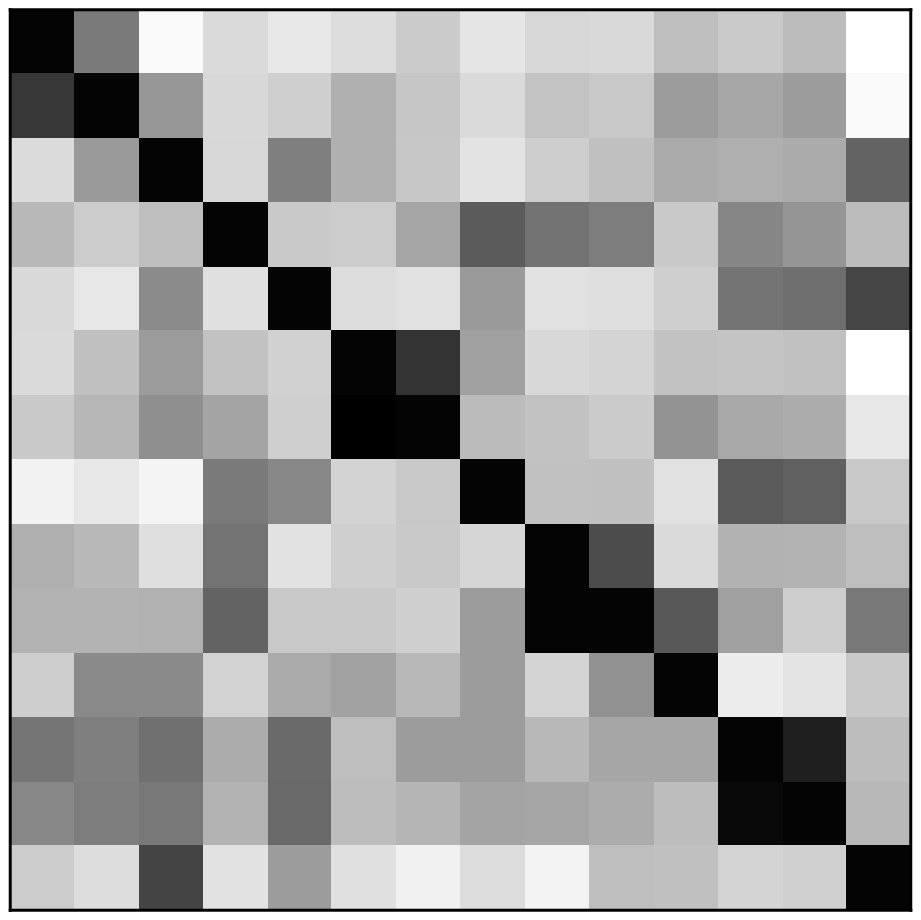}}
  \subfigure[correlation coefficient]{\includegraphics[width=.3\textwidth]{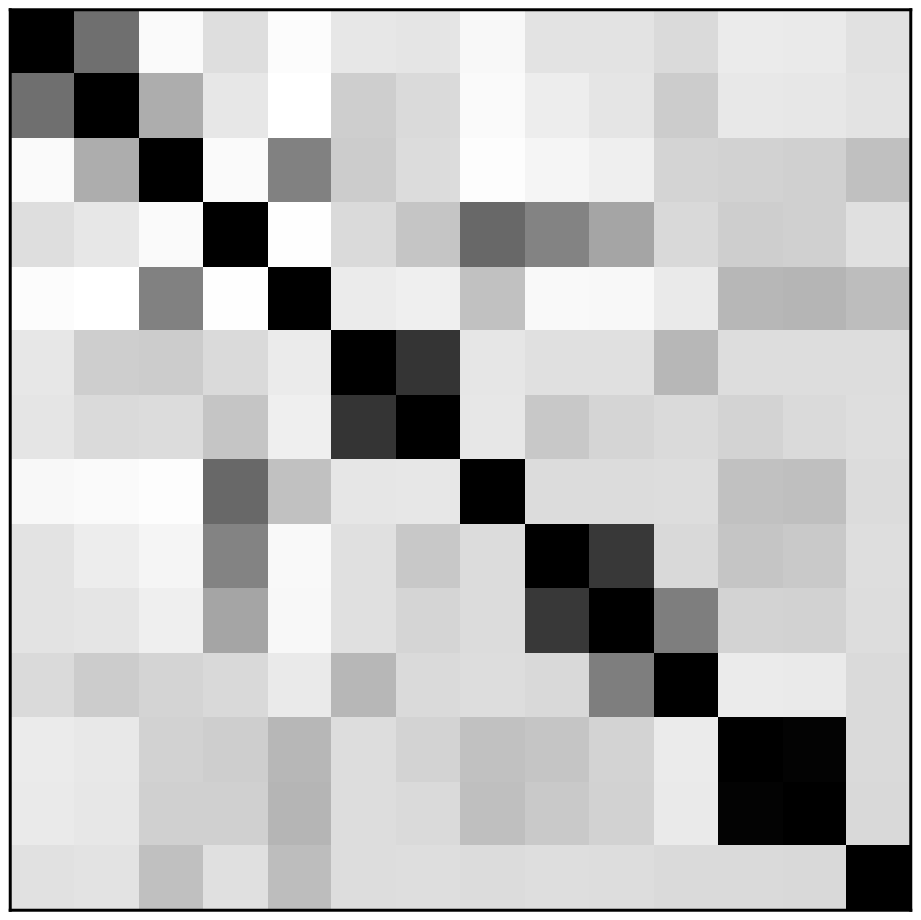}}
\caption{
  The trained memory transformation matrix $\mathbf{W}$ with SRN and the
  correlation coefficient of the yeast data set.
  Each cell represents the correlation between two labels.
  Each row of the memory weight is normalized for the diagonal element to be
  1 so it can be compared with correlation coefficient.
  }
\label{fig:mem_weights}
\end{figure}

Figure \ref{fig:mem_weights} plots the learned memory transformation matrix and the
correlation coefficient of the labels.
We can clearly see that RethinkNet is able to capture the label correlation
information,
although we also found that such result in some data set can be noisy.
The finding suggests that $\mathbf{W}$ may carry not only label correlation
but also other data set factors.
For example, the RNN model may learn that the prediction of a certain label
does not come with a high accuracy.
Therefore, even if another label is highly correlated with this one, the model
would not give it a high weight.

\subsection{Cost-Sensitive Reweighted Loss Function}
Cost information is another important piece of information that should be
considered when solving an MLC problem.
Different cost function values each label differently, so we should set
the importance of each label differently.
One way to encode such property is to weight each label in the loss function
according to its importance.
The problem would become how to estimate the label importance.

The difference between a label predicted correctly and incorrectly under
the cost function can be used to estimate the importance of the label.
To evaluate the importance of a single label, knowing other
labels is required for most costs.
We leverage the sequential nature of RethinkNet where temporary predictions
are made between each of the iterations.
Using the temporary prediction to fill out all other labels, we will be
able to estimate the importance of each label.

The weight of each label is designed as equation \eqref{eq:weigthing}.
For $t=1$, where no prior prediction exists, the labels are set with equal
importance.
For $t>1$, we use $\mathbf{\hat{y}}^{(t)}_n[i]_0$ and $\mathbf{\hat{y}}^{(t)}_n[i]_1$
to represent the label vector $\mathbf{\hat{y}}^{(t)}_n$ when the $i$-th label
is set to $0$ and $1$ respectively.
The weight of the $i$-th label is the cost difference between
$\mathbf{\hat{y}}^{(t)}_n[i]_0$ and $\mathbf{\hat{y}}^{(t)}_n[i]_1$.
This weighting approach can be used to estimate the effect of each label
under current prediction with the given cost function.
Such method echos the design of CFT \citep{CL2014b}.
\begin{equation} \label{eq:weigthing}
\begin{split}
\mathbf{w}^{(1)}_n[i] = 1, \quad
\mathbf{w}^{(t)}_n[i] = |C(\mathbf{y}_n, \mathbf{\hat{y}}^{(t-1)}_n[i]_0) -
                          C(\mathbf{y}_n, \mathbf{\hat{y}}^{(t-1)}_n[i]_1)|
\end{split}
\end{equation}

In usual MLC setting, the loss function adopted for neural network is binary
cross-entropy.
To accept the weight in loss function, we formulated the weighted binary
cross-entropy as Equation \eqref{eq:weigthed-bin-entropy}.
For $t=1$, the weight for all labels are set to 1 since there is no
prediction to reference.
For $t=2, \ldots K$, the weights are updated using the previous prediction.
Note that when the given cost function is Hamming loss, the labels in each
iteration are weighted the same and the weighting is reduced to the same as
in BR.
\begin{equation} \label{eq:weigthed-bin-entropy}
    \frac{1}{N} \sum^{N}_{n=1} \sum^{B}_{t=1} \sum^{K}_{i=1}
        - \mathbf{w}^{(t)}_n[i] (
          \mathbf{y}_n[i] \log p(\mathbf{\hat{y}}^{(t)}_n[i])
          + (1 - \mathbf{y}_n[i]) \log (1 - p(\mathbf{\hat{y}}^{(t)}_n[i])))
\end{equation}

\begin{table}[h]
  \scriptsize
  \caption{Comparison between MLC algorithms.}
  \label{tab:algo-comp}
  \centering
  \begin{tabular}{l C{.25\textwidth} cc}
    \toprule
algorithm  & memory content       & cost-sensitivity & feature extraction \\
\hline
BR         &  -                         & -          & -                  \\
CC         & former prediction          & -          & -                  \\
CC-DP      & optimal ordered prediction & -          & -                  \\
PCC        & former prediction          & v          & -                  \\
CFT        & former prediction          & v          & -                  \\
CNN-RNN    & former prediction in RNN   & -          & CNN                \\
Att-RNN    & former prediction in RNN   & -          & CNN + attention    \\
RethinkNet & full prediction in RNN     & v          & general NN         \\
    \bottomrule
  \end{tabular}
\end{table}
Table \ref{tab:algo-comp} shows a comparison between MLC algorithms.
RethinkNet is able to consider both the label correlation and the cost
information.
Its structure allows it to be extended easily with other neural network for
advance feature extraction, so it can be easily adopted to solve image labeling
problems.
In Section \ref{experiments}, we will demonstrate that these advantages can be
turned into better performance.

\section{Experiments} \label{experiments}

The experiments are evaluated on 11 real-world data sets \citep{mulan}.
The data set statistics are shown in Table \ref{tab:mlc-ds}.
The data set is split with 75\% training and 25\% testing randomly
and the feature vectors are scaled to $[0, 1]$.
Experiments are repeated 10 times with the mean and standard error (ste)
of the testing loss/score recorded.
The results are evaluated with Hamming loss, Rank loss, F1 score, Accuracy
score \citep{CL2014b}.
We use $\downarrow$ to indicate the lower value for the criterion is
better and $\uparrow$ to indicate the higher value is better.

RethinkNet is implemented 
with \textsc{tensorflow} \citep{tensorflow2015-whitepaper}. 
The RNN layer can be interchanged with different variations of RNN
including SRN, LSTM GRU and IRNN.
A 25\% dropout on the memory matrix of RNN is added.
A single fully-connected layer is used for the dense layer and Nesterov Adam
(Nadam)~\citep{dozat2016incorporating} is used to optimize the model.
The model is trained until converges or reach $1,000$ epochs and the batch
size is fixed to $256$.
We added L2 regularization to the training parameters and the regularization
strength is search within $(10^{-8}, \ldots, 10^{-1})$ with three-fold
cross-validation.
The implementation of RethinkNet can be found here \footnote{\url{https://github.com/yangarbiter/multilabel-learn}}.

\begin{table}[h]
  \scriptsize
  \caption{Statistics on multi-label data sets}
  \label{tab:mlc-ds}
  \centering
  \begin{tabular}{lccccc}
    \toprule
    data set        & feature dim.  & label dim. & data points  & cardinality & density \\
    \midrule
    emotions       & 72            & 6          &  593    & 1.869  & 0.311	\\
    scene          & 2407          & 6          &  2407   & 1.074  & 0.179 \\
    yeast          & 2417          & 14         &  2417   & 4.237  & 0.303 \\
    birds          & 260           & 19         &  645    & 1.014	 & 0.053	\\
    tmc2007        & 500           & 22         &  28596  & 2.158  & 0.098 \\
    arts1          & 23146         & 26         &  7484   & 1.654  & 0.064 \\
    medical        & 120           & 45         &  978    & 1.245  & 0.028 \\
    enron          & 1001          & 53         &  1702   & 3.378  & 0.064 \\
    bibtex         & 1836          & 159        &  7395   & 2.402  & 0.015 \\	
    CAL500         & 68            & 174        &  502    & 26.044 & 0.150 \\
    Corel5k        & 499           & 374        &  5000   & 3.522	 & 0.009 \\
    \bottomrule
  \end{tabular}
\end{table}

\subsection{Rethinking} \label{exp:rethinking}

\begin{figure*}[ht!]
  \centering
  \subfigure[scene]{\includegraphics[width=.85\textwidth]{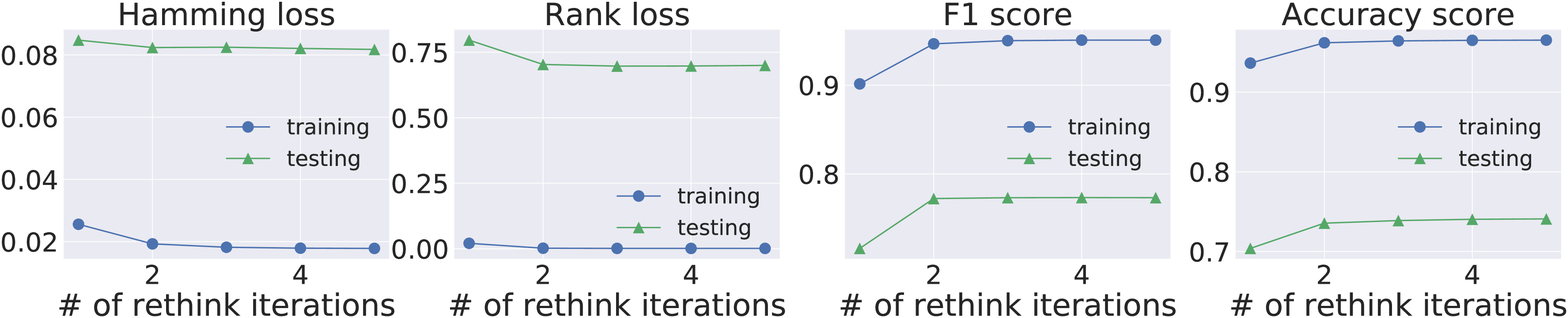}} \\
  \subfigure[yeast]{\includegraphics[width=.85\textwidth]{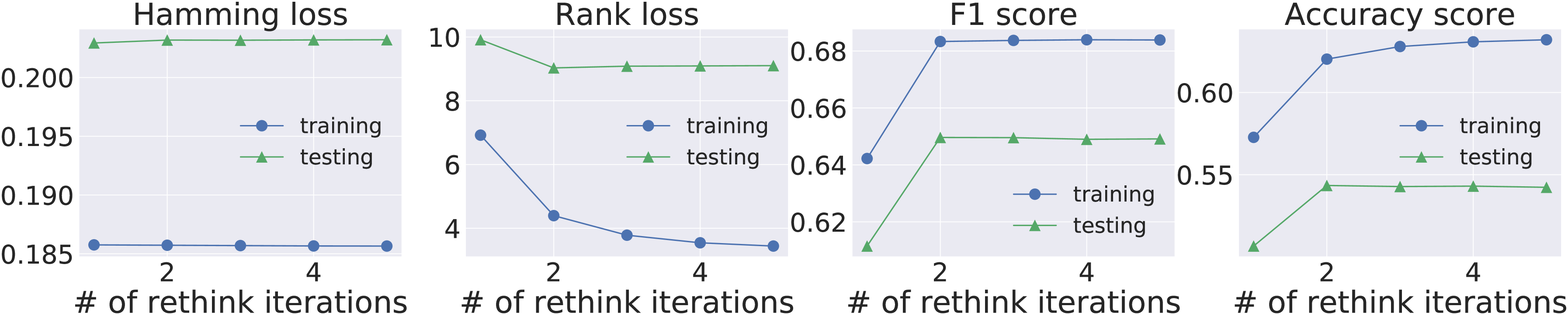}} \\
  \subfigure[medical]{\includegraphics[width=.85\textwidth]{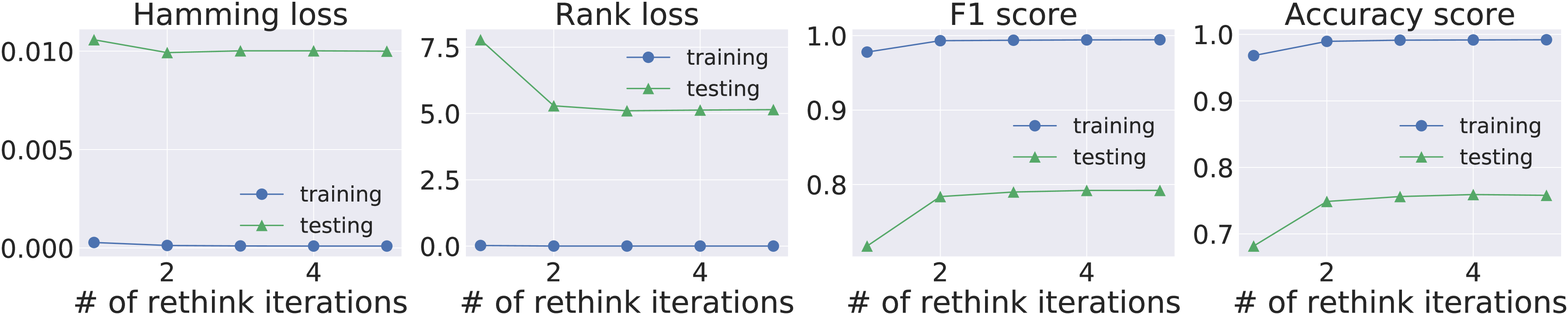}} \\
  \subfigure[CAL500]{\includegraphics[width=.85\textwidth]{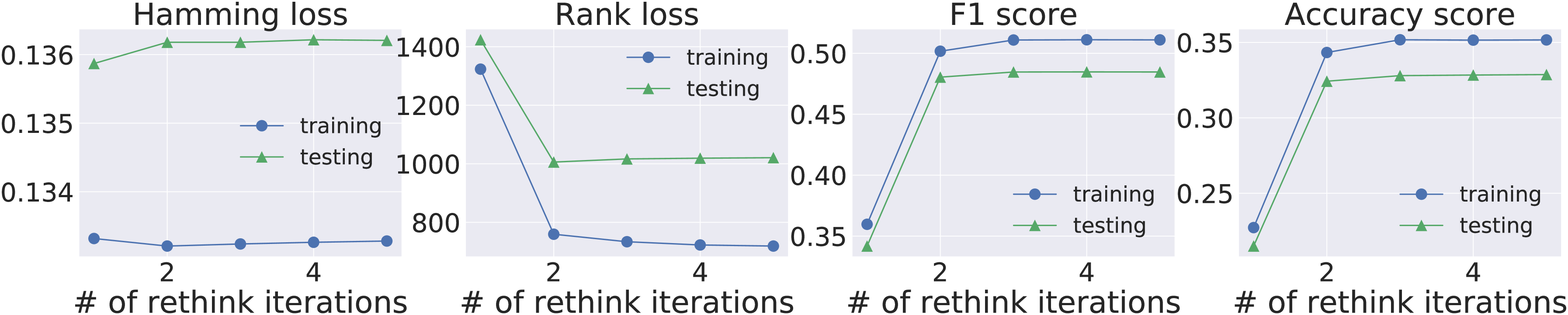}}
\caption{The average performance versus number of rethink iteration.}
\label{fig:rethinking}
\end{figure*}

In Section \ref{proposed-model}, we claim that RethinkNet is able to improve
through iterations of rethinking.
We justify our claim with this experiment.
In this experiment, we use the simplest form of RNN, SRN, in the RNN layer of
RethinkNet and the dimensionality of the RNN layer is fixed to $128$.
We set the number of rethink iterations $B=5$ and plot the training and
testing loss/score on Figure \ref{fig:rethinking}.

From the figure, we can see that for cost functions like Rank loss, F1 score,
Accuracy score, which relies more on utilizing label correlation, achieved
significant improvement over the increase of rethink iteration.
Hamming loss is a criterion that evaluates each label independently and
algorithms that does not consider label correlation like BR perform well
on such criterion \citep{read2009classifier}.
The result shows that the performance generally converges at around
the third rethink iteration.
For efficiency, the rest of experiments will be fixed with $B=3$.

\begin{figure}
  \center
  \includegraphics[width=.45\textwidth]{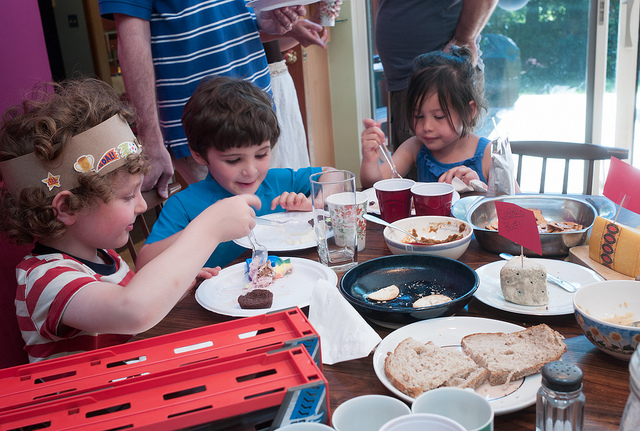}
  \caption{An example from the MSCOCO data set \citep{lin2014microsoft}
  with ground truth labels
  'person', 'cup', 'fork', 'knife', 'spoon', 'bowl', 'cake', 'chair', 'dining
  table'.
  }
  \label{fig:mscoco_example}
\end{figure}

To further observe the rethinking process, we also trained RethinkNet on the
MSCOCO \citep{lin2014microsoft} data set and observe its behavior on
real-world images.
The detailed experimental setup can be found in Section \ref{img_dataset}.
Take Figure \ref{fig:mscoco_example} as example.
In the first iteration, RethinkNet predict label 'person', 'cup', 'fork',
'bowl', 'chair', 'dining table' exists in the figure.
These are labels that are easier to detect.
Using the knowledge this may be a scene on a dining table, the probability
that there exist 'knife' or 'spoon' should be increased. 
In the second iteration, RethinkNet further predicted that 'bottle', 'knife',
'spoon' are also in the example.
In the third iteration, RethinkNet found that the bottle should not be in the
figure and exclude it from the prediction.

\subsection{Effect of Reweighting}

We conducted this experiment to verify the cost-sensitive reweighting can
really use the cost information to reach a better performance.
The performance of RethinkNet with and without reweighting under
Rank loss, F1 score and Accuracy score is compared.
Hamming loss is the same before and after reweighting so it is not shown in
the result.
Table \ref{tab:reweight} lists the mean and standard error (ste) of each
experiment and it demonstrates that on almost all data sets, reweighting the
loss function for RethinkNet yields better result.

\begin{table*}[ht!]
  \scriptsize
  \caption{Experimental results (mean $\pm$ ste) of the performance in Rank loss ($\downarrow$),  F1 score ($\uparrow$),  Accuracy score ($\uparrow$) of
  non-reweighted and reweighted RethinkNet (best ones are bold) }
  \label{tab:reweight}
  \centering
  \begin{tabular}{l|cc|cc|cc}
    \toprule
 & \multicolumn{2}{c}{Rank loss} & \multicolumn{2}{c}{F1 score} & \multicolumn{2}{c}{Accuracy score} \\ 
\midrule 
data set & non-reweighted & reweighted & non-reweighted & reweighted & non-reweighted & reweighted \\ 
\midrule 
emotions & $ 3.48 \pm .13 $& $ \mathbf{1.82 \pm .25 }$& $ .652 \pm .012 $& $ \mathbf{.687 \pm .006 }$& $ .574 \pm .007 $& $ \mathbf{.588 \pm .007 }$ \\ 
scene & $ 2.50 \pm .03 $& $ \mathbf{.72 \pm .01 }$& $ .750 \pm .005 $& $ \mathbf{.772 \pm .005 }$& $ .721 \pm .008 $& $ \mathbf{.734 \pm .005 }$ \\ 
yeast & $ 13.3 \pm .05 $& $ \mathbf{9.04 \pm .09 }$& $ .612 \pm .005 $& $ \mathbf{.648 \pm .004 }$& $ .500 \pm .005 $& $ \mathbf{.538 \pm .004 }$ \\ 
birds & $ 8.21 \pm .43 $& $ \mathbf{4.24 \pm .32 }$& $ \mathbf{.237 \pm .011 }$& $ .236 \pm .012 $& $ \mathbf{.195 \pm .013 }$& $ .193 \pm .008 $ \\ 
tmc2007 & $ 9.59 \pm .32 $& $ \mathbf{5.37 \pm .02 }$& $ \mathbf{.754 \pm .004 }$& $ .748 \pm .009 $& $ \mathbf{.691 \pm .003 }$& $ .690 \pm .002 $ \\ 
Arts1 & $ 19.6 \pm .05 $& $ \mathbf{13.0 \pm .2 }$& $ .351 \pm .005 $& $ \mathbf{.365 \pm .003 }$& $ .304 \pm .005 $& $ \mathbf{.315 \pm .004 }$ \\ 
medical & $ 27.2 \pm .2 $& $ \mathbf{5.6 \pm .2 }$& $ .793 \pm .006 $& $ \mathbf{.795 \pm .004 }$& $ \mathbf{.761 \pm .006 }$& $ .760 \pm .006 $ \\ 
enron & $ 60.3 \pm 2.5 $& $ \mathbf{39.7 \pm .5 }$& $ .544 \pm .007 $& $ \mathbf{.604 \pm .007 }$& $ .436 \pm .006 $& $ \mathbf{.480 \pm .004 }$ \\ 
Corel5k & $ 654. \pm 1. $& $ \mathbf{524. \pm 2. }$& $ .169 \pm .002 $& $ \mathbf{.257 \pm .001 }$& $ .118 \pm .001 $& $ \mathbf{.164 \pm .002 }$ \\ 
CAL500 & $ 1545. \pm 17. $& $ \mathbf{997. \pm 12. }$& $ .363 \pm .003 $& $ \mathbf{.484 \pm .003 }$& $ .231 \pm .005 $& $ \mathbf{.328 \pm .002 }$ \\ 
bibtex & $ 186. \pm 1. $& $ \mathbf{115. \pm 1. }$& $ .390 \pm .002 $& $ \mathbf{.398 \pm .002 }$& $ .320 \pm .002 $& $ \mathbf{.329 \pm .002 }$ \\ 
    \bottomrule
  \end{tabular}
\end{table*}

\subsection{Compare with Other MLC Algorithms}

We compare RethinkNet with other state-of-the-art MLC and CSMLC
algorithms in this experiment.
The competing algorithms includes the binary relevance (BR), probabilistic
classifier chain (PCC), classifier chain (CC), dynamic programming based
classifier chain (CC-DP), condensed filter tree (CFT).
To compare with the RNN structure used in CNN-RNN~\citep{wang2016cnn},
we implemented a classifier chains using RNN (CC-RNN) as competitor.
CC-RNN is CNN-RNN without the CNN layer since we are dealing with general
data sets.
BR is implemented using a feed-forward neural network with a $128$ neurons hidden layer.
We coupled both CC-RNN and RethinkNet with a $128$ neurons LSTM.
CC-RNN and BR are trained using same approach as RethinkNet.
Training $K$ independent feed-forward neural network is too computationally
heavy, so we coupled CFT, PCC, CC with L2-regularized logistic regression.
CC-DP is coupled with SVM as it is derived for it.
We adopt the implementation from \textsc{scikit-learn}~\citep{scikit-learn}
for both the L2-regularized logistic regression and linear SVM.
The regularization strength for these models are searched within $(10^{-4},
10^{-3}, \ldots, 10^{4})$ with three-fold cross-validation.
PCC does not have inference rule derived for Accuracy score and we use the F1
score inference rule as an alternative in view of the similarity in the
formula.
Other parameters not mentioned are kept with default of the implementation.

The experimental results are shown on Table \ref{tab:mlc-comps} and
the t-test results are on Table \ref{tab:ttest}.
Note that we cannot get the result of CC-DP in two weeks on the data sets
Corel5k, CAL500 and bibtex so they are not listed.
In terms of average ranking and t-test results, RethinkNet yields a
superior performance.
On Hamming loss, all algorithms are generally competitive.
For Rank loss, F1 score and Accuracy score, CSMLC algorithms (RethinkNet,
PCC, CFT) begin to take the lead.
Even the parameters of cost-insensitive algorithms are tuned on the target
evaluation criteria, they are not able to compete with cost-sensitive
algorithms.
This demonstrates the importance of developing cost-sensitive algorithms.

All three CSMLC algorithms has similar performance on Rank loss and
RethinkNet performs slightly better on F1 score.
For Accuracy score, since PCC is not able to directly utilize the cost
information of, this makes PCC performs slightly poor.

When comparing with deep structures (RethinkNet, CC-RNN, BR), only BR is
competitive under Hamming loss with RethinkNet.
On all other settings, RethinkNet is able to outperform the other two
competitors.
CC-RNN learns an RNN with sequence length being the number of labels ($K$).
When $K$ gets large, the depth of CC-RNN can go very deep making it hard to
train with fixed learning rate in our setting and failed to perform well on
these data sets.
This demonstrates that RethinkNet is a better designed deep structure to
solve CSMLC problem.


\begin{table}[ht!]
  \scriptsize
  \caption{RethinkNet versus the competitors based on t-test at 95\%
  confidence level }
  \label{tab:ttest}
  \centering
  \begin{tabular}{lccccccc}
    \toprule
 (\#win/\#tie/\#loss) & PCC & CFT & CC-DP & CC & CC-RNN & BR \\ 
\midrule 
hamming  $(\downarrow)$ & 6/1/4 & 3/4/4 & 5/2/1 & 6/1/4 & 8/3/0 & 3/6/2 \\ 
rank loss $(\downarrow)$ & 5/1/5 & 5/2/4 & 7/1/0 & 10/1/0 & 10/1/0 & 10/1/0 \\ 
f1       $(\uparrow)$   & 6/2/3 & 5/4/2 & 5/2/1 & 8/3/0 & 10/1/0 & 9/2/0 \\ 
acc      $(\uparrow)$   & 7/1/3 & 5/4/2 & 5/1/2 & 7/4/0 & 9/2/0 & 9/2/0 \\ 
\midrule 
total    & 24/5/15 & 18/14/12 & 22/6/4 & 31/9/4 & 37/7/0 & 31/11/2 \\
\bottomrule
  \end{tabular}
\end{table}

\begin{table}[ht!]
  \scriptsize
  \caption{Experimental results on MSCOCO data set.}
  \label{tab:image_comp}
  \centering
  \begin{tabular}{lcccccc}
    \toprule
 & baseline & CNN-RNN & Att-RNN & RethinkNet & \\ 
\midrule 
hamming  $(\downarrow)$ & $0.0279 $ & $0.0267$ & $0.0270 $ & $\mathbf{0.0234}  $ \\
rank loss $(\downarrow)$ & $60.4092$ & $56.6088$ & $43.5248$ & $\mathbf{35.2552}$ \\
f1       $(\uparrow)$ & $0.5374 $ & $0.5759$   & $0.6309 $ & $\mathbf{0.6622}  $ \\
acc      $(\uparrow)$ & $0.4469 $ & $0.4912$   & $0.5248 $ & $\mathbf{0.5724}  $\\
\bottomrule
  \end{tabular}
\end{table}

\begin{table*}[ht!]
  \tiny
  \caption{Experimental results (mean $\pm$ ste) on different criteria
  (best results in bold)}
  \label{tab:mlc-comps}
  \centering
  \begin{tabular}{lccccccc}
    \toprule
\multicolumn{8}{c}{Hamming loss $\downarrow$} \\
\midrule 
data set & RethinkNet & PCC & CFT & CC-DP & CC & CC-RNN & BR \\ 
\midrule 
emotions  & $ .191 \pm .005 [2] $ & $ .219 \pm .005 [6] $ & $ .194 \pm .003 [4] $ & $ .213 \pm .004 [5] $ & $ .219 \pm .005 [7] $ & $ .192 \pm .004 [3] $ & $ \mathbf{.190 \pm .004 [1]} $ \\ 
scene  & $ \mathbf{.081 \pm .001 [1]} $ & $ .101 \pm .001 [5] $ & $ .095 \pm .001 [4] $ & $ .104 \pm .002 [7] $ & $ .101 \pm .001 [6] $ & $ .087 \pm .001 [2] $ & $ .087 \pm .003 [2] $ \\ 
yeast  & $ \mathbf{.205 \pm .001 [1]} $ & $ .218 \pm .001 [6] $ & $ \mathbf{.205 \pm .002 [1]} $ & $ .214 \pm .002 [4] $ & $ .218 \pm .001 [7] $ & $ .215 \pm .002 [5] $ & $ .205 \pm .002 [2] $ \\ 
birds  & $ \mathbf{.048 \pm .001 [1]} $ & $ .050 \pm .001 [3] $ & $ .051 \pm .001 [6] $ & $ .050 \pm .002 [3] $ & $ .050 \pm .001 [3] $ & $ .053 \pm .002 [7] $ & $ .049 \pm .001 [2] $ \\ 
tmc2007  & $ \mathbf{.046 \pm .000 [1]} $ & $ .058 \pm .000 [5] $ & $ .057 \pm .000 [4] $ & $ .058 \pm .000 [5] $ & $ .058 \pm .000 [5] $ & $ .047 \pm .001 [2] $ & $ .048 \pm .000 [3] $ \\ 
Arts1  & $ .062 \pm .001 [5] $ & $ .060 \pm .000 [2] $ & $ .060 \pm .000 [2] $ & $ .065 \pm .001 [6] $ & $ .060 \pm .000 [2] $ & $ .068 \pm .001 [7] $ & $ \mathbf{.058 \pm .000 [1]} $ \\ 
medical  & $ .010 \pm .000 [1] $ & $ .010 \pm .000 [1] $ & $ .011 \pm .000 [5] $ & $ \mathbf{.010 \pm .000 [1]} $ & $ .010 \pm .000 [1] $ & $ .023 \pm .000 [7] $ & $ .011 \pm .000 [6] $ \\ 
enron  & $ .047 \pm .000 [4] $ & $ .046 \pm .000 [1] $ & $ \mathbf{.046 \pm .000 [1]} $ & $ .047 \pm .000 [4] $ & $ .046 \pm .000 [1] $ & $ .059 \pm .000 [7] $ & $ .048 \pm .000 [6] $ \\ 
Corel5k  & $ .009 \pm .000 [1] $ & $ \mathbf{.009 \pm .000 [1]} $ & $ .009 \pm .000 [1] $ & $ - \pm - $ & $ .009 \pm .000 [1] $ & $ .009 \pm .000 [1] $ & $ .009 \pm .000 [1] $ \\ 
CAL500  & $ \mathbf{.137 \pm .001 [1]} $ & $ .138 \pm .001 [3] $ & $ .138 \pm .001 [3] $ & $ - \pm - $ & $ .138 \pm .001 [3] $ & $ .149 \pm .001 [6] $ & $ .137 \pm .001 [1] $ \\ 
bibtex  & $ .013 \pm .000 [2] $ & $ .013 \pm .000 [2] $ & $ .013 \pm .000 [2] $ & $ - \pm - $ & $ .013 \pm .000 [2] $ & $ .015 \pm .000 [6] $ & $ \mathbf{.012 \pm .000 [1]} $ \\ 
\bottomrule
avg. rank & $\mathbf{1.82}$ & $3.18$	& $3.00$ & $4.38$ &	$3.45$ & $4.82$ & $2.36$ \\
\bottomrule

\multicolumn{8}{c}{Rank loss $\downarrow$} \\
\midrule 
data set & RethinkNet & PCC & CFT & CC-DP & CC & CC-RNN & BR \\ 
\midrule 
emotions  & $ \mathbf{1.48 \pm .04 [1]} $ & $ 1.63 \pm .05 [3] $ & $ 1.59 \pm .03 [2] $ & $ 3.64 \pm .02 [4] $ & $ 3.64 \pm .02 [4] $ & $ 3.64 \pm .02 [4] $ & $ 3.64 \pm .02 [4] $ \\
scene  & $ \mathbf{.72 \pm .01 [1]} $ & $ .88 \pm .03 [2] $ & $ .96 \pm .04 [3] $ & $ 2.59 \pm .01 [5] $ & $ 2.61 \pm .01 [6] $ & $ 2.49 \pm .04 [4] $ & $ 2.61 \pm .01 [6] $ \\
yeast  & $ 8.89 \pm .11 [2] $ & $ 9.76 \pm .08 [3] $ & $ \mathbf{8.83 \pm .09 [1]} $ & $ 13.23 \pm .04 [5] $ & $ 13.16 \pm .07 [4] $ & $ 19.47 \pm .04 [7] $ & $ 13.23 \pm .04 [5] $ \\
birds  & $ \mathbf{4.32 \pm .27 [1]} $ & $ 4.66 \pm .18 [2] $ & $ 4.90 \pm .20 [3] $ & $ 8.51 \pm .28 [4] $ & $ 8.51 \pm .28 [4] $ & $ 8.51 \pm .28 [4] $ & $ 8.51 \pm .28 [4] $ \\
tmc2007  & $ 5.22 \pm .04 [3] $ & $ 4.32 \pm .01 [2] $ & $ \mathbf{3.89 \pm .01 [1]} $ & $ 12.32 \pm .03 [6] $ & $ 12.14 \pm .03 [5] $ & $ 21.44 \pm .02 [7] $ & $ 11.39 \pm .04 [4] $ \\
Arts1  & $ 13.0 \pm .1 [3] $ & $ \mathbf{12.2 \pm .1 [1]} $ & $ 12.9 \pm .0 [2] $ & $ 19.7 \pm .0 [4] $ & $ 19.7 \pm .0 [4] $ & $ 19.7 \pm .0 [4] $ & $ 19.7 \pm .0 [4] $ \\
medical  & $ 5.3 \pm .1 [2] $ & $ \mathbf{4.4 \pm .1 [1]} $ & $ 6.0 \pm .2 [3] $ & $ 27.2 \pm .1 [4] $ & $ 27.3 \pm .1 [5] $ & $ 27.3 \pm .1 [5] $ & $ 27.3 \pm .1 [5] $ \\
enron  & $ \mathbf{40.1 \pm .6 [1]} $ & $ 42.8 \pm .6 [3] $ & $ 42.2 \pm .6 [2] $ & $ 49.0 \pm .5 [5] $ & $ 48.7 \pm .5 [4] $ & $ 82.3 \pm .5 [7] $ & $ 52.8 \pm .4 [6] $ \\
Corel5k  & $ 527. \pm 2. [3] $ & $ \mathbf{426. \pm 1. [1]} $ & $ 460. \pm 2. [2] $ & $ - \pm - $ & $ 654. \pm 1. [5] $ & $ 653. \pm 1. [4] $ & $ 654. \pm 1. [5] $ \\
CAL500  & $ \mathbf{1040. \pm 8. [1]} $ & $ 1389. \pm 10. [3] $ & $ 1234. \pm 10. [2] $ & $ - \pm - $ & $ 1599. \pm 13. [5] $ & $ 1915. \pm 10. [6] $ & $ 1568. \pm 9. [4] $ \\
bibtex  & $ 114. \pm 1. [3] $ & $ \mathbf{99. \pm 1. [1]} $ & $ 112. \pm 1. [2] $ & $ - \pm - $ & $ 186. \pm 1. [4] $ & $ 186. \pm 1. [4] $ & $ 186. \pm 1. [4] $ \\
\bottomrule
avg. rank & $\mathbf{1.91}$ & $2$ & $2.09$ & $4.63$ & $4.55$ & $5.09$ & $4.64$ \\
\bottomrule

\multicolumn{8}{c}{F1 score $\uparrow$} \\
\midrule 
data set & RethinkNet & PCC & CFT & CC-DP & CC & CC-RNN & BR \\ 
\midrule 
emotions  & $ \mathbf{.690 \pm .007 [1]} $ & $ .654 \pm .004 [3] $ & $ .655 \pm .006 [2] $ & $ .616 \pm .008 [7] $ & $ .620 \pm .008 [6] $ & $ .649 \pm .007 [4] $ & $ .639 \pm .009 [5] $ \\ 
scene  & $ \mathbf{.765 \pm .003 [1]} $ & $ .734 \pm .004 [3] $ & $ .730 \pm .003 [5] $ & $ .711 \pm .005 [6] $ & $ .710 \pm .004 [7] $ & $ .742 \pm .004 [2] $ & $ .731 \pm .006 [4] $ \\ 
yeast  & $ \mathbf{.651 \pm .003 [1]} $ & $ .598 \pm .003 [4] $ & $ .646 \pm .003 [2] $ & $ .617 \pm .003 [3] $ & $ .587 \pm .003 [6] $ & $ .577 \pm .007 [7] $ & $ .593 \pm .012 [5] $ \\ 
birds  & $ .235 \pm .016 [2] $ & $ \mathbf{.251 \pm .011 [1]} $ & $ .217 \pm .009 [5] $ & $ .208 \pm .008 [6] $ & $ .225 \pm .008 [3] $ & $ .087 \pm .006 [7] $ & $ .221 \pm .008 [4] $ \\ 
tmc2007  & $ \mathbf{.765 \pm .002 [1]} $ & $ .683 \pm .001 [6] $ & $ .718 \pm .001 [4] $ & $ .676 \pm .001 [7] $ & $ .684 \pm .001 [5] $ & $ .732 \pm .009 [3] $ & $ .740 \pm .001 [2] $ \\ 
Arts1  & $ .385 \pm .006 [3] $ & $ \mathbf{.425 \pm .002 [1]} $ & $ .411 \pm .003 [2] $ & $ .375 \pm .003 [4] $ & $ .365 \pm .004 [5] $ & $ .076 \pm .002 [7] $ & $ .359 \pm .003 [6] $ \\ 
medical  & $ .790 \pm .004 [3] $ & $ \mathbf{.812 \pm .004 [1]} $ & $ .780 \pm .006 [4] $ & $ .799 \pm .004 [2] $ & $ .778 \pm .007 [5] $ & $ .333 \pm .010 [7] $ & $ .755 \pm .006 [6] $ \\ 
enron  & $ \mathbf{.601 \pm .003 [1]} $ & $ .557 \pm .002 [3] $ & $ .599 \pm .004 [2] $ & $ .539 \pm .004 [6] $ & $ .556 \pm .004 [4] $ & $ .424 \pm .011 [7] $ & $ .548 \pm .004 [5] $ \\ 
Corel5k  & $ .232 \pm .003 [3] $ & $ .233 \pm .001 [2] $ & $ \mathbf{.259 \pm .001 [1]} $ & $ - \pm - $ & $ .156 \pm .002 [5] $ & $ .000 \pm .000 [6] $ & $ .164 \pm .001 [4] $ \\ 
CAL500  & $ \mathbf{.485 \pm .002 [1]} $ & $ .405 \pm .002 [3] $ & $ .477 \pm .002 [2] $ & $ - \pm - $ & $ .347 \pm .003 [5] $ & $ .048 \pm .001 [6] $ & $ .360 \pm .004 [4] $ \\ 
bibtex  & $ .394 \pm .005 [2] $ & $ \mathbf{.425 \pm .002 [1]} $ & $ .393 \pm .003 [3] $ & $ - \pm - $ & $ .393 \pm .002 [4] $ & $ .000 \pm .000 [6] $ & $ .385 \pm .003 [5] $ \\ 
\bottomrule
avg. rank & $\mathbf{1.73}$ & $2.55$ &	$2.91$ & $5.13$ &	$5.00$ & $5.64$ &	$4.55$ \\
\bottomrule

\multicolumn{8}{c}{Accuracy score $\uparrow$} \\
\midrule 
data set & RethinkNet & PCC & CFT & CC-DP & CC & CC-RNN & BR \\ 
\midrule 
emotions  & $ \mathbf{.600 \pm .007 [1]} $ & $ .556 \pm .006 [4] $ & $ .566 \pm .006 [3] $ & $ .534 \pm .008 [7] $ & $ .538 \pm .008 [6] $ & $ .568 \pm .006 [2] $ & $ .545 \pm .009 [5] $ \\ 
scene  & $ \mathbf{.737 \pm .003 [1]} $ & $ .693 \pm .005 [7] $ & $ .700 \pm .004 [4] $ & $ .699 \pm .005 [5] $ & $ .699 \pm .004 [6] $ & $ .718 \pm .006 [2] $ & $ .707 \pm .008 [3] $ \\ 
yeast  & $ \mathbf{.541 \pm .004 [1]} $ & $ .482 \pm .003 [7] $ & $ .533 \pm .003 [2] $ & $ .514 \pm .003 [3] $ & $ .486 \pm .003 [5] $ & $ .484 \pm .008 [6] $ & $ .495 \pm .008 [4] $ \\ 
birds  & $ .205 \pm .009 [7] $ & $ .211 \pm .009 [6] $ & $ .592 \pm .010 [3] $ & $ \mathbf{.596 \pm .009 [1]} $ & $ .592 \pm .010 [2] $ & $ .525 \pm .013 [5] $ & $ .589 \pm .011 [4] $ \\ 
tmc2007  & $ \mathbf{.700 \pm .003 [1]} $ & $ .578 \pm .001 [7] $ & $ .618 \pm .001 [4] $ & $ .587 \pm .001 [6] $ & $ .595 \pm .001 [5] $ & $ .634 \pm .033 [3] $ & $ .667 \pm .002 [2] $ \\ 
Arts1  & $ .320 \pm .003 [5] $ & $ .351 \pm .002 [2] $ & $ \mathbf{.370 \pm .003 [1]} $ & $ .337 \pm .003 [3] $ & $ .326 \pm .003 [4] $ & $ .071 \pm .002 [7] $ & $ .308 \pm .002 [6] $ \\ 
medical  & $ .754 \pm .004 [3] $ & $ \mathbf{.780 \pm .004 [1]} $ & $ .751 \pm .006 [4] $ & $ .771 \pm .004 [2] $ & $ .750 \pm .008 [5] $ & $ .304 \pm .007 [7] $ & $ .728 \pm .008 [6] $ \\ 
enron  & $ \mathbf{.482 \pm .003 [1]} $ & $ .429 \pm .004 [6] $ & $ .480 \pm .004 [2] $ & $ .437 \pm .004 [5] $ & $ .452 \pm .004 [3] $ & $ .324 \pm .011 [7] $ & $ .441 \pm .004 [4] $ \\ 
Corel5k  & $ .161 \pm .002 [2] $ & $ .148 \pm .001 [3] $ & $ \mathbf{.168 \pm .001 [1]} $ & $ - \pm - $ & $ .111 \pm .001 [5] $ & $ .000 \pm .000 [6] $ & $ .113 \pm .001 [4] $ \\ 
CAL500  & $ \mathbf{.326 \pm .001 [1]} $ & $ .255 \pm .001 [3] $ & $ .320 \pm .001 [2] $ & $ - \pm - $ & $ .218 \pm .002 [5] $ & $ .027 \pm .002 [6] $ & $ .230 \pm .003 [4] $ \\ 
bibtex  & $ .327 \pm .002 [4] $ & $ \mathbf{.353 \pm .002 [1]} $ & $ .328 \pm .002 [2] $ & $ - \pm - $ & $ .327 \pm .002 [3] $ & $ .000 \pm .000 [6] $ & $ .326 \pm .002 [5] $ \\ 
\bottomrule
avg. rank & $\mathbf{2.45}$ & $4.27$ & $2.55$ & $4.00$ & $4.45$ & $5.18$ & $4.27$ \\
\bottomrule

  \end{tabular}
\end{table*}

\subsection{Comparison on Image Data Set}\label{img_dataset}

The CNN-RNN and Att-RNN algorithms are designed to solve image labeling
problems.
The purpose of this experiment is to understand how RethinkNet 
performs on such task comparing with CNN-RNN and Att-RNN.
We use the data set MSCOCO \citep{lin2014microsoft} and the training testing
split provided by them.
Pre-trained Resnet-50 \citep{DBLP:journals/corr/HeZRS15} is adopted for
feature extraction.
The competing models include
logistic regression as baseline,
CNN-RNN,
Att-RNN,
and RethinkNet.
The implementation of Att-RNN is from the original author and
other models are implemented with \textsc{keras}.
The models are fine tuned with the pre-trained Resnet-50.
The results on testing data set are shown on Table \ref{tab:image_comp}.
The result shows that RethinkNet is able to outperform state-of-the-art
deep learning models that are designed for image labeling.

\subsection{Using Different Variations of RNN}

In this experiment, we compare the performance of RethinkNet using different
forms of RNN on the RNN layer in RethinkNet.
The competitors includes SRN, LSTM, GRU, and IRNN.
We tuned the label embedding dimensionality so that the total number of
trainable parameters are around $200,000$ for each form of RNN.
The results are evaluated on two more commonly seen cost functions, Rank loss
and F1 score, and shown on Table \ref{tab:comp-rnn}.

Different variations of RNN differs in the way they manipulate the memory.
In terms of testing result, we can see that SRN and LSTM are two better choices.
GRU and IRNN tends to overfit too much causing their testing performance
to drop.
Among SRN and LSTM, SRN tends to have a slightly larger discrepancy between
training and testing performance.
We can also observed that many data sets performs better with the same
variation of RNN across cost functions.
This indicates that different data set may require different form of memory
manipulation.

\begin{table*}[ht!]
  \tiny
  \caption{Experimental results with different RNN for RethinkNet.
  Evaluated in Rank loss $\downarrow$ and F1 score $\uparrow$ (best results are in bold)}
  \label{tab:comp-rnn}
  \centering
  \begin{tabular}{l|cc|cc|cc|cc}
    \toprule
\multicolumn{9}{c}{Rank loss $\downarrow$} \\
\midrule 
data set & \multicolumn{2}{c}{SRN} & \multicolumn{2}{c}{GRU} & \multicolumn{2}{c}{LSTM} & \multicolumn{2}{c}{IRNN} \\ 
 & training & testing & training & testing & training & testing & training & testing\\ 
\midrule 
emotions & $ 1.06 \pm .51 $ & $ 1.81 \pm .26 $ & $ .45 \pm .13 $ & $ 1.54 \pm .04 $ & $ .68 \pm .06 $ & $ \mathbf{1.50 \pm .06} $ & $ .00 \pm .00 $ & $ 1.60 \pm .04 $  \\ 
scene & $ .001 \pm .000 $ & $ \mathbf{.706 \pm .012} $ & $ .001 \pm .001 $ & $ .708 \pm .015 $ & $ .002 \pm .000 $ & $ .715 \pm .006 $ & $ .001 \pm .000 $ & $ .763 \pm .005 $  \\ 
yeast & $ .34 \pm .05 $ & $ 9.69 \pm .10 $ & $ .70 \pm .06 $ & $ 9.93 \pm .16 $ & $ 3.67 \pm 1.09 $ & $ \mathbf{9.18 \pm .16} $ & $ .01 \pm .01 $ & $ 10.17 \pm .10 $  \\ 
birds & $ .02 \pm .01 $ & $ 4.29 \pm .28 $ & $ .00 \pm .00 $ & $ 4.44 \pm .31 $ & $ .01 \pm .01 $ & $ \mathbf{4.25 \pm .28} $ & $ .00 \pm .00 $ & $ 4.34 \pm .30 $  \\ 
tmc2007 & $ .11 \pm .03 $ & $ \mathbf{5.01 \pm .07} $ & $ .12 \pm .03 $ & $ 5.25 \pm .04 $ & $ .11 \pm .05 $ & $ 5.17 \pm .07 $ & $ .07 \pm .01 $ & $ 5.13 \pm .05 $  \\ 
Arts1 & $ .7 \pm .1 $ & $ 13.3 \pm .2 $ & $ 5.8 \pm .2 $ & $ 13.1 \pm .2 $ & $ 5.5 \pm .0 $ & $ \mathbf{13.0 \pm .1} $ & $ .2 \pm .0 $ & $ 14.3 \pm .2 $  \\ 
medical & $ .00 \pm .00 $ & $ \mathbf{4.75 \pm .22} $ & $ .00 \pm .00 $ & $ 5.85 \pm .27 $ & $ .01 \pm .00 $ & $ 5.40 \pm .35 $ & $ .00 \pm .00 $ & $ 6.13 \pm .42 $  \\ 
enron & $ .4 \pm .0 $ & $ 39.7 \pm .4 $ & $ .4 \pm .0 $ & $ 39.1 \pm .6 $ & $ .4 \pm .0 $ & $ \mathbf{38.8 \pm .5} $ & $ .4 \pm .0 $ & $ 39.0 \pm .5 $  \\ 
Corel5k & $ 0. \pm 0. $ & $ \mathbf{524. \pm 2.} $ & $ 0. \pm 0. $ & $ 532. \pm 2. $ & $ 0. \pm 0. $ & $ 526. \pm 2. $ & $ 0. \pm 0. $ & $ 534. \pm 2. $  \\ 
CAL500 & $ 893. \pm 124. $ & $ \mathbf{1035. \pm 21.} $ & $ 377. \pm 11. $ & $ 1101. \pm 13. $ & $ 544. \pm 16. $ & $ 1053. \pm 11. $ & $ 8. \pm 1. $ & $ 1358. \pm 14. $  \\ 
bibtex & $ 0. \pm 0. $ & $ 117. \pm 1. $ & $ 0. \pm 0. $ & $ 121. \pm 2. $ & $ 0. \pm 0. $ & $ 122. \pm 1. $ & $ 0. \pm 0. $ & $ \mathbf{109. \pm 3.} $  \\ 
    \toprule

\multicolumn{9}{c}{F1 score $\uparrow$} \\
\midrule 
data set & \multicolumn{2}{c}{SRN} & \multicolumn{2}{c}{GRU} & \multicolumn{2}{c}{LSTM} & \multicolumn{2}{c}{IRNN} \\ 
 & training & testing & training & testing & training & testing & training & testing\\ 
\midrule 
emotions & $ .794 \pm .023 $ & $ .682 \pm .010 $ & $ .811 \pm .022 $ & $ .680 \pm .003 $ & $ .788 \pm .004 $ & $ \mathbf{.690 \pm .006} $ & $ .836 \pm .051 $ & $ .681 \pm .008 $  \\ 
scene & $ .919 \pm .002 $ & $ \mathbf{.769 \pm .003} $ & $ .961 \pm .034 $ & $ .757 \pm .003 $ & $ .857 \pm .025 $ & $ .753 \pm .011 $ & $ .931 \pm .027 $ & $ .764 \pm .004 $  \\ 
yeast & $ .724 \pm .027 $ & $ .641 \pm .005 $ & $ .687 \pm .008 $ & $ .643 \pm .005 $ & $ .709 \pm .002 $ & $ \mathbf{.651 \pm .003} $ & $ .691 \pm .022 $ & $ .640 \pm .004 $  \\ 
birds & $ .513 \pm .020 $ & $ .235 \pm .014 $ & $ .546 \pm .008 $ & $ .243 \pm .015 $ & $ .508 \pm .014 $ & $ .240 \pm .013 $ & $ .552 \pm .006 $ & $ \mathbf{.248 \pm .015} $  \\ 
tmc2007 & $ .990 \pm .002 $ & $ \mathbf{.771 \pm .003} $ & $ .991 \pm .002 $ & $ .764 \pm .003 $ & $ .982 \pm .004 $ & $ .758 \pm .004 $ & $ .983 \pm .003 $ & $ .757 \pm .001 $  \\ 
Arts1 & $ .763 \pm .009 $ & $ \mathbf{.364 \pm .005} $ & $ .395 \pm .026 $ & $ .323 \pm .003 $ & $ .406 \pm .033 $ & $ .320 \pm .004 $ & $ .522 \pm .090 $ & $ .344 \pm .009 $  \\ 
medical & $ .995 \pm .001 $ & $ \mathbf{.793 \pm .005} $ & $ .988 \pm .000 $ & $ .792 \pm .002 $ & $ .976 \pm .001 $ & $ .791 \pm .006 $ & $ .999 \pm .000 $ & $ .786 \pm .009 $  \\ 
enron & $ .689 \pm .004 $ & $ .605 \pm .003 $ & $ .695 \pm .003 $ & $ \mathbf{.610 \pm .003} $ & $ .665 \pm .003 $ & $ .603 \pm .003 $ & $ .740 \pm .008 $ & $ .608 \pm .007 $  \\ 
Corel5k & $ .340 \pm .002 $ & $ \mathbf{.260 \pm .002} $ & $ .325 \pm .002 $ & $ .255 \pm .003 $ & $ .475 \pm .018 $ & $ .230 \pm .005 $ & $ .409 \pm .016 $ & $ .221 \pm .009 $  \\ 
CAL500 & $ .491 \pm .006 $ & $ .474 \pm .004 $ & $ .507 \pm .002 $ & $ .483 \pm .004 $ & $ .506 \pm .001 $ & $ \mathbf{.485 \pm .002} $ & $ .493 \pm .002 $ & $ .478 \pm .001 $  \\ 
bibtex & $ .995 \pm .001 $ & $ .391 \pm .005 $ & $ .860 \pm .050 $ & $ .385 \pm .004 $ & $ .854 \pm .006 $ & $ .379 \pm .003 $ & $ .928 \pm .022 $ & $ \mathbf{.399 \pm .003} $  \\ 
    \bottomrule
  \end{tabular}
\end{table*}

\section{Conclusion}\label{conclusion}
Classic multi-label classification (MLC) algorithms predict labels as a
sequence to model the label correlation.
However, these approaches face the problem of ordering the labels in the
sequence.
In this paper, we reformulate the sequence prediction problem to avoid the
issue.
By mimicking the human rethinking process, we propose a novel cost-sensitive
multi-label classification (CSMLC) algorithm called RethinkNet.
RethinkNet takes the process of gradually polishing its prediction as the
sequence to predict.
We adopt the recurrent neural network (RNN) to predict the sequence, and the
memory in the RNN can then be used to store the label correlation information.
In addition, we also modified the loss function to take in the cost
information, and thus make RethinkNet cost-sensitive.
Extensive experiments demonstrate that RethinkNet is able to
outperform other MLC and CSMLC algorithms on general data sets.
On image data set, RethinkNet is also able to exceed state-of-the-art
image labeling algorithms in performance.
The results suggest that RethinkNet is a promising algorithm for solving
CSMLC using neural network.





\bibliography{rethinknet}

\end{document}